%% file: main.tex
\ifdefined\pdfoutput
\ifdefined\pdfmajorversion
\pdfoutput=1
\fi
\fi
\documentclass[11pt]{article}
\usepackage[utf8]{inputenc}
\usepackage[T1]{fontenc}
\usepackage{geometry}
\usepackage{graphicx}
\usepackage{booktabs}
\usepackage{array}
\usepackage{tabularx}
\usepackage{url}
\usepackage{float}
\usepackage{enumitem}
\setlist{nosep}
\usepackage[hidelinks]{hyperref}
\graphicspath{{./}}
\newcolumntype{Y}{>{\raggedright\arraybackslash}X}
\newcolumntype{C}[1]{>{\centering\arraybackslash}p{#1}}
\title{Skill-Augmented AI Agents for Medical Research Analysis: An Exploratory Multi-Model Human Evaluation in an NSCLC Transcriptomic Biomarker Task}
\author{
Qianyu Yao$^{1,*}$, Fei Sun$^{1,*}$, Bocheng Huang$^{1,*}$, Wei Chen$^{1}$,\\
Jiarui Jiang$^{1}$, Shu Quan$^{1}$, Yifei Chen$^{1}$, Weijie Xu$^{1}$,\\
Bo Li$^{1}$, Liping Su$^{1}$, Ruoqiong Wu$^{1}$, Huhai Hong$^{1}$, Huimei Wang$^{1,\dagger}$\\[0.75em]
\small $^{1}$AIPOCH PTE. LTD., Singapore\\
\small $^{*}$These authors contributed equally as co-first authors.\\
\small $^{\dagger}$Corresponding author: Huimei Wang.
}
\date{}
\begin{document}
\maketitle
\begin{abstract}
\setlength{\parindent}{0pt}
\noindent\textbf{Background.} Large language models and AI agents are increasingly used to support biomedical research, but native model outputs may omit key analytical steps, misuse methods, or overstate conclusions. We evaluated whether autonomous access to a medical research skill package was associated with higher-quality AI-generated transcriptomic research-analysis outputs compared with native AI without skills.

\noindent\textbf{Methods.} We conducted an exploratory multi-model human evaluation using an NSCLC immunotherapy biomarker task. Six model backbones were tested. The evaluation included 21 anonymized outputs: 9 native-AI outputs and 12 skill-augmented outputs generated through an AI agent implementation represented by OpenClaw. Four non-expert biomedical reviewers and two blinded experts evaluated each output, with two ratings from each reviewer type. The primary outcome was expert-rated overall quality.

\noindent\textbf{Results.} Skill-augmented outputs showed directionally higher expert overall quality than native-AI outputs (mean 5.50 vs 5.11; difference=0.39; bootstrap 95\% CI, -0.04 to 0.90; Welch p=0.156). Non-expert reviewer quality showed the same direction (mean 4.72 vs 4.47; difference=0.26; bootstrap 95\% CI, -0.25 to 0.80; Welch p=0.373). Expert agreement was limited (single-rating ICC=-0.15), and model-specific effects were descriptive and heterogeneous.

\noindent\textbf{Conclusions.} Autonomous skill access showed a directional quality signal in this exploratory sample, but the signal was smaller than expert-rating noise and should not be interpreted as confirmatory evidence. The findings primarily motivate larger evaluations of skill-augmented AI agents with stronger reliability controls, platform replication, and biological-validity assessment.
\end{abstract}

\section{Introduction}

Large language models (LLMs) and artificial intelligence (AI) agents are increasingly being explored as assistants for biomedical research, including literature interpretation, study planning, statistical reasoning, and data analysis. Prior work has shown that large foundation models can encode clinically relevant knowledge and perform competitively on medical question-answering tasks, while broader discussions of generalist medical AI have highlighted their potential to support complex biomedical workflows [1,2]. However, native LLM outputs remain vulnerable to omissions, inappropriate method selection, weak validation logic, and overconfident claims.

Tool use and agentic orchestration have emerged as strategies for improving reliability. Benchmarks and systems for application programming interface (API) and tool use, such as API-Bank [3], Gorilla [4], ToolLLM [5], and TaskBench [6], emphasize that model performance should be evaluated not only by language generation but also by whether a model can select and execute appropriate tools. More recent skill-based agent research has shifted attention toward reusable skills, skill routing, dependency-aware retrieval, and workflow-level benchmarking [7-12]. However, less is known about whether reusable medical research skills improve downstream human-rated research-analysis quality across different model backbones, especially in biomedical workflows where evidence selection, endpoint design, validation logic, and claim calibration must be integrated.

Medical research is a demanding domain for skill orchestration because a usable output must connect evidence, protocol design, dataset selection, endpoint definition, preprocessing, statistical analysis, validation, biological interpretation, and explicit limitations. Transcriptomic biomarker research is an appropriate stress test: it requires public dataset selection, cohort design, expression matrix preprocessing, differential expression analysis, pathway enrichment, immune microenvironment analysis, feature selection, modeling, and validation. In non-small cell lung cancer (NSCLC) immunotherapy research, candidate biomarkers must also be interpreted cautiously because sample size, endpoint mismatch, overfitting, and limited biological plausibility can substantially weaken downstream claims.

This study evaluated whether autonomous access to a medical research skill package was associated with higher-rated AI-generated research-analysis outputs compared with native AI alone. The contribution is threefold: first, it provides a human-evaluated biomedical use case for skill-augmented AI agents; second, it distinguishes downstream research-analysis quality from conventional tool-use or routing accuracy benchmarks; and third, it examines descriptive model-specific heterogeneity while explicitly treating the findings as hypothesis-generating.

\section{Methods}

\subsection{Study Design}

This was an exploratory, multi-model, human-evaluated comparison between native AI and a skill-augmented AI agent. The native-AI strategy used model outputs generated without access to medical research skills. The skill-augmented strategy used an AI agent with access to a medical research skill package and autonomous skill routing. OpenClaw was used as the representative AI agent platform for the skill-augmented implementation. The study assessed the quality of generated research-analysis outputs rather than final manuscripts or clinically validated biomarkers.

\subsection{Model Backbones}

Six model backbones were included: GPT-5.4, Claude Sonnet 4.6, GLM-5.1, DeepSeek-V4 Pro, Kimi K2.6, and MiniMax-M2.7. Models were selected pragmatically to cover widely used frontier or near-frontier systems available to the study team across multiple providers and deployment ecosystems; they were not intended to represent a systematic sample of all available LLMs. The final evaluation dataset included 21 outputs: 9 native-AI outputs and 12 skill-augmented outputs. Repeated outputs generated during the final evaluation process were retained to reflect the real-world variability of agentic workflows.

\subsection{Generation Protocol}

All model backbones were evaluated on the same task prompt and expected-output requirements. The native-AI strategy received the task without access to the medical research skill package. The skill-augmented strategy used the same task in an AI agent environment with autonomous access to the skill package. Outputs were retained if they produced reviewable research-analysis material, including complete reports, stage summaries, or output files. Partial or nonstandard outputs were not excluded if they remained reviewable, because the study aimed to evaluate real-world agentic output quality rather than only ideal completed runs. Proprietary model decoding parameters were not uniformly exposed by the execution environments; therefore, the analysis reports the evaluated model backbones and generated outputs rather than claiming exact low-level reproducibility of model sampling settings. The supplement provides the exact task prompt, output inclusion rules, evaluated output counts (Table~\ref{tab:supp-output-counts}), rating anchors (Table~\ref{tab:supp-rating-anchors}), and analysis artifacts (Table~\ref{tab:supp-analysis-artifacts}).

\subsection{Unified Research Task}

All models received the same research task: to use public transcriptomic data to construct a multi-gene signature for predicting immunotherapy response in NSCLC and to explore the role of programmed cell death (PCD) mechanisms, including ferroptosis, cuproptosis, and pyroptosis, in immunotherapy resistance. The expected output was a research-analysis plan and data-analysis workflow with a level of complexity comparable to a journal article with an impact factor (IF) of approximately 5 in the corresponding year. This phrase was used to set task complexity and expected completeness; it was not intended to predict journal acceptance, actual publication quality, or the future impact factor of any journal. Required components included public dataset selection, cohort design, endpoint definition, preprocessing, differential expression analysis, candidate gene screening, model construction, validation strategy, immune microenvironment analysis, mechanistic interpretation, key figures and tables, and manual review points.

\subsection{Skill Package}

The medical research skill package was sourced from the \texttt{awesome-med-research-skills} repository in the \href{https://github.com/aipoch/medical-research-skills/tree/main/awesome-med-research-skills}{AIPOCH medical research skills collection}. It included evidence-insight, protocol-design, and data-analysis skills. The evaluated skills were a mixture of procedural guidance modules and execution-oriented analysis modules. Data-analysis capabilities included expression matrix normalization, batch effect correction, differential expression analysis, Gene Ontology/Kyoto Encyclopedia of Genes and Genomes (GO/KEGG) enrichment, Gene Set Enrichment Analysis (GSEA), Gene Set Variation Analysis (GSVA), CIBERSORT or single-sample GSEA (ssGSEA) immune infiltration analysis, Estimation of STromal and Immune cells in MAlignant Tumours using Expression data (ESTIMATE) scoring, least absolute shrinkage and selection operator (LASSO) or elastic-net feature selection, machine-learning modeling, receiver operating characteristic (ROC) analysis, external validation, calibration, decision curve analysis, survival analysis where applicable, protein-protein interaction (PPI) networks, transcription factor (TF)-target networks, competing endogenous RNA (ceRNA) or long non-coding RNA (lncRNA) networks where data allowed, clustering, dimensionality reduction, sample correlation analysis, and weighted gene co-expression network analysis (WGCNA) where statistically appropriate. These modules should be understood as broad bioinformatics and biomedical-research workflow modules rather than disease- or mechanism-specific modules for NSCLC programmed cell death. In the executed benchmark workflow, the downloaded public datasets were checked for lung-cancer or NSCLC relevance before downstream analysis, but dedicated ferroptosis-, cuproptosis-, or pyroptosis-specific gene-set curation was not independently validated within the deployed skill environment. Therefore, programmed-cell-death interpretation depended on the model's retrieval, reasoning, and use of available general enrichment or pathway-analysis skills. Academic writing skills were not part of the evaluated task. The study did not verify whether each generated output's proposed programmed-cell-death gene sets or biomarkers were biologically valid.

\subsection{Output Anonymization}

All outputs were anonymized before review. Model names, model families, generation strategy labels, platform configuration, and obvious experimental cues were removed. Each output received a random anonymous identifier. The mapping between anonymous IDs, models, and generation strategies was stored separately and was not available to reviewers.

\subsection{Human Evaluation}

Four non-expert biomedical reviewers evaluated clarity, completeness, perceived credibility, usability, workflow coherence, and perceived risk. These reviewers had biomedical or medical-research familiarity but were not treated as specialist NSCLC transcriptomics or biostatistics adjudicators. Each output received two non-expert ratings. The non-expert quality score was calculated as the mean of 7-point Likert-scale items covering content clarity, methodological completeness, perceived credibility, feasibility, and practical usability, with higher scores indicating better perceived quality. The non-expert workflow score was calculated from items assessing skill selection, step sequencing, upstream-downstream continuity, and integration of multiple analytical components. Non-expert risk perception was calculated from items assessing potential methodological error, statistical overinterpretation, inappropriate skill use, and the need for human review; higher scores indicated greater perceived risk. For quality and workflow items, scale anchors were 1 = very poor or strongly disagree, 4 = neutral or partially acceptable, and 7 = excellent or strongly agree. For perceived-risk items, 1 indicated very low perceived risk and 7 indicated very high perceived risk.

Two blinded experts independently reviewed all outputs. Expert ratings used 7-point Likert-scale items covering research question clarity, evidence and research gap, objectives, cohort and endpoint design, transcriptomic preprocessing, differential expression and enrichment strategy, immune analysis, biomarker construction, modeling and validation, statistical and bioinformatic appropriateness, workflow integration, risk and limitation statements, feasibility, and overall quality. The primary outcome was expert-rated overall quality. Reviewers assessed anonymized outputs without access to model names, generation strategies, or platform configuration.

\subsection{Statistical Analysis}

The primary unit of analysis was the anonymized output after averaging the two ratings available for each reviewer type. Scores were summarized by generation strategy using means, standard deviations, medians, quartiles, and ranges. Strategy-level comparisons between native-AI and skill-augmented outputs were evaluated using Welch t tests and Mann-Whitney U tests, with nonparametric bootstrap 95\% confidence intervals (CIs) calculated for mean skill-minus-native differences. Model-level skill-minus-native differences were calculated by averaging outputs within each model-strategy combination and then subtracting the model's native-AI score from its skill-augmented score. These model-level contrasts were treated as descriptive only because several cells contained one native-AI output and two skill-augmented outputs. Repeated outputs were retained as separate generated artifacts, but model-level summaries were used to reduce overinterpretation of repeated-run dependence. Inter-rater agreement was assessed using two-way absolute-agreement intraclass correlation coefficients (ICCs) for expert ratings and exploratory one-way random-effects ICCs for non-expert ratings because non-expert ratings were not fully crossed by reviewer. No multiplicity adjustment was applied across the five strategy-level outcome summaries or the six model-level descriptive contrasts. All analyses were exploratory.

\section{Results}

\subsection{Dataset and Quality Control}

\begin{figure}[htbp]
\centering
\includegraphics[width=0.95\linewidth]{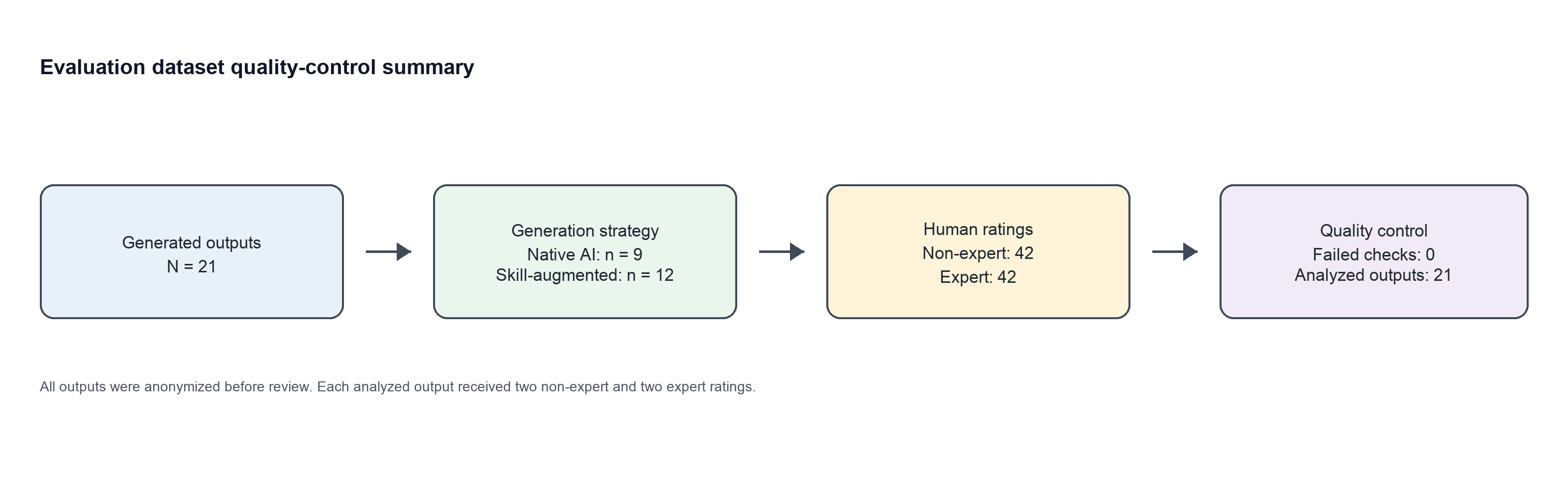}
\caption{Evaluation dataset quality-control flow. The figure summarizes the included anonymized outputs, generation-strategy distribution, human-rating volume, and non-expert review quality-control checks.}
\label{fig:qc-flow}
\end{figure}

The evaluation dataset included 21 anonymized outputs: 9 native-AI outputs and 12 skill-augmented outputs. Each output received two non-expert-review ratings and two expert-review ratings. No non-expert-review record failed attention or reading-confirmation checks (Figure~\ref{fig:qc-flow}).

\subsection{Overall Comparison by Generation Strategy}

Expert overall quality was directionally higher for skill-augmented outputs than for native-AI outputs (mean 5.50 vs 5.11; skill-minus-native difference=0.39; bootstrap 95\% CI, -0.04 to 0.90; Cohen's d=0.73; Welch p=0.156; Mann-Whitney p=0.150). Non-expert reviewer quality showed the same direction (mean 4.72 vs 4.47; difference=0.26; bootstrap 95\% CI, -0.25 to 0.80; Welch p=0.373; Mann-Whitney p=0.355). Expert methodological quality showed a larger difference favoring skill augmentation (difference=0.47; bootstrap 95\% CI, 0.12 to 0.95), whereas non-expert perceived risk was lower on average but uncertain (difference=-0.30; bootstrap 95\% CI, -1.02 to 0.42).

\begin{table}[htbp]
\centering
\caption{Quality scores by generation strategy.}
\small
\begin{tabularx}{\linewidth}{Y C{1.7cm} C{2.0cm} C{1.9cm} C{2.0cm}}
\toprule
Outcome & Native AI mean (n=9) & Skill-augmented mean (n=12) & Skill-minus-native difference & Bootstrap 95\% CI \\
\midrule
Expert overall quality & 5.11 & 5.50 & 0.39 & -0.04 to 0.90 \\
Expert methodological quality & 4.79 & 5.26 & 0.47 & 0.12 to 0.95 \\
Non-expert reviewer quality & 4.47 & 4.72 & 0.26 & -0.25 to 0.80 \\
Non-expert workflow score & 4.71 & 5.03 & 0.32 & -0.10 to 0.70 \\
Non-expert perceived risk & 4.93 & 4.63 & -0.30 & -1.02 to 0.42 \\
\bottomrule
\end{tabularx}
\end{table}

\begin{figure}[htbp]
\centering
\includegraphics[width=0.95\linewidth]{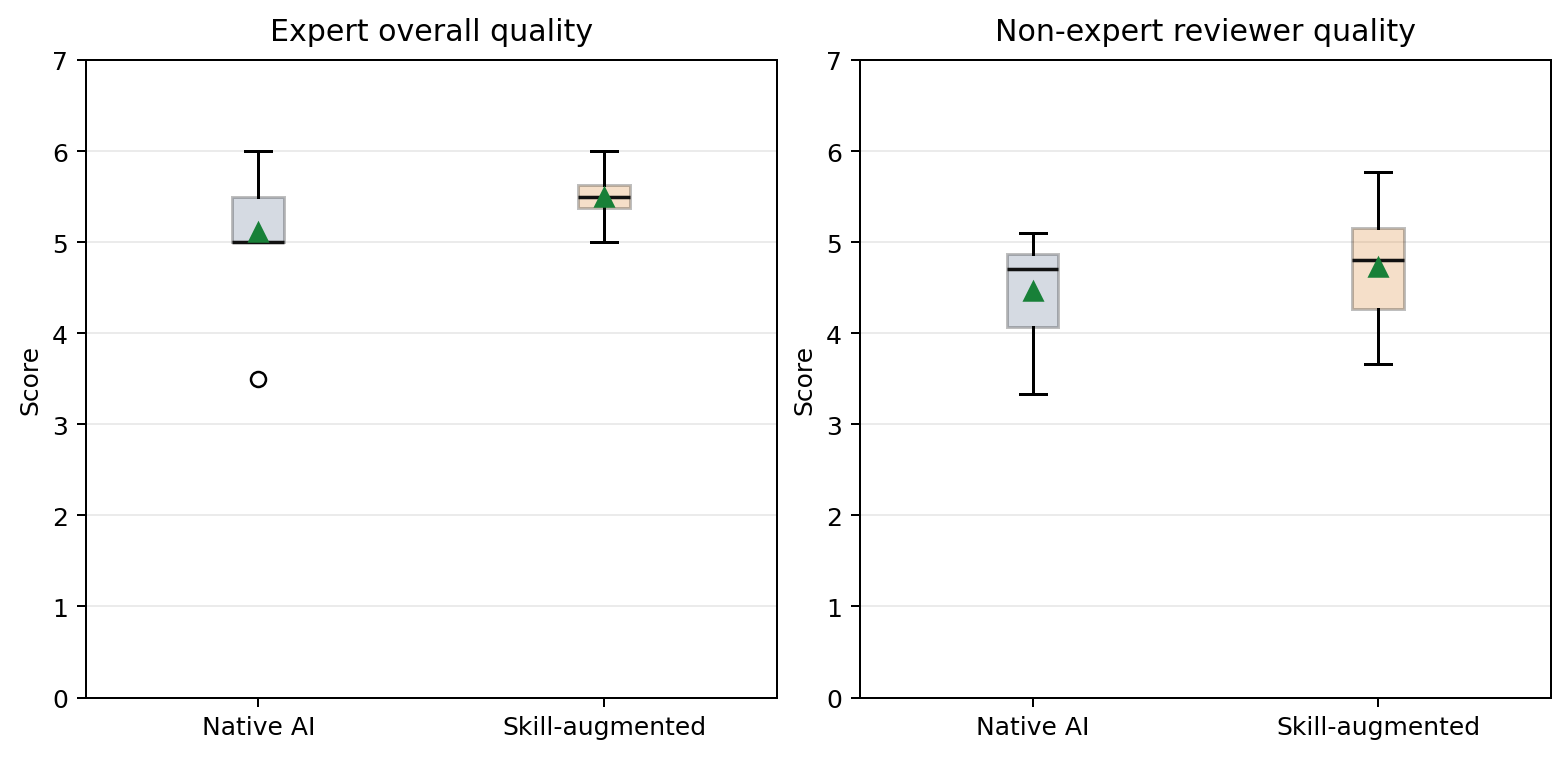}
\caption{Overall quality scores by generation strategy. Boxplots show expert-rated overall quality and non-expert reviewer quality for native-AI and skill-augmented outputs. Boxes indicate interquartile ranges, horizontal lines indicate medians, whiskers indicate non-outlier ranges, circles indicate outliers, and green triangles indicate means.}
\label{fig:overall-quality}
\end{figure}

\begin{figure}[htbp]
\centering
\includegraphics[width=0.95\linewidth]{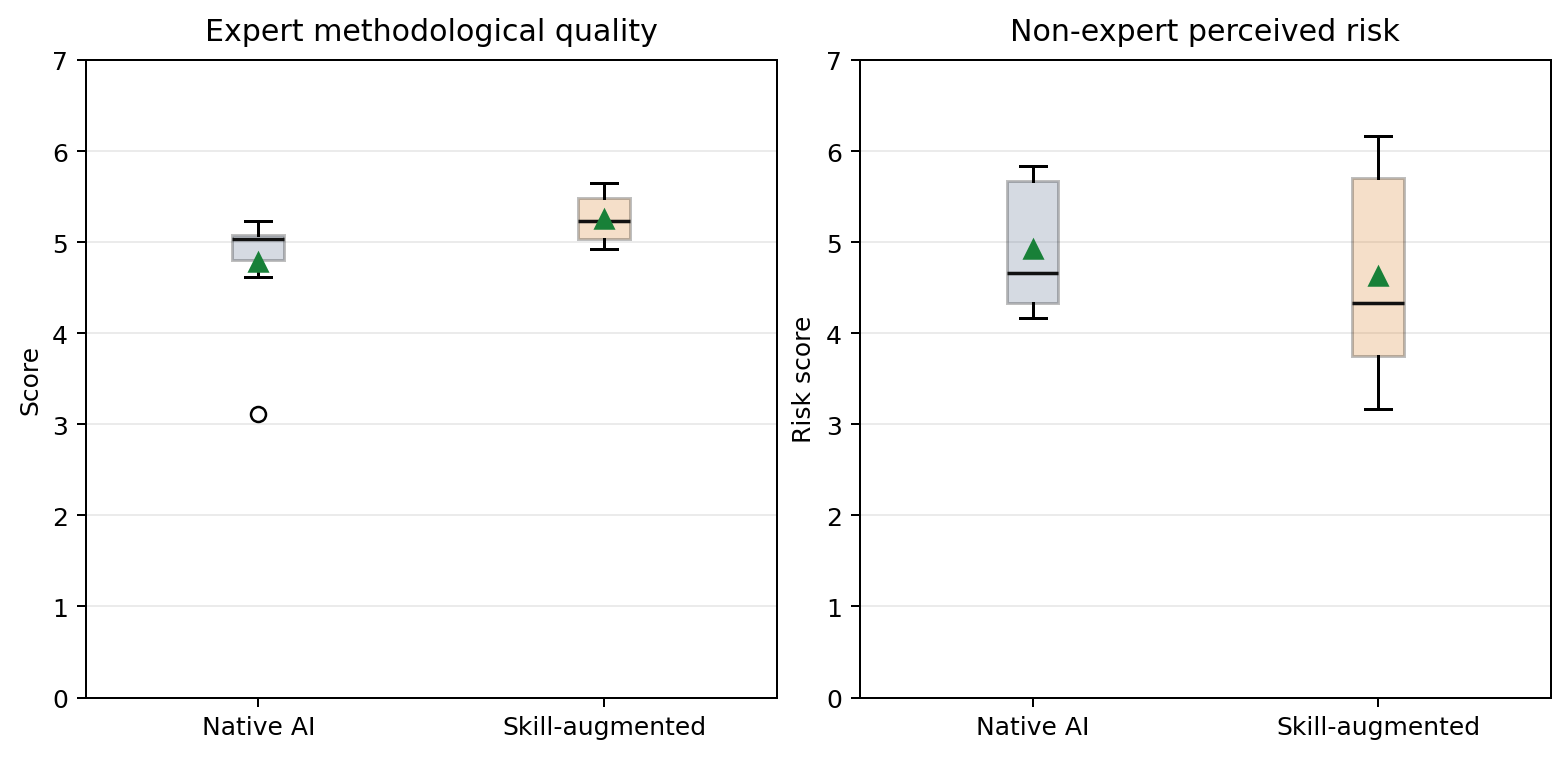}
\caption{Expert methodological quality and non-expert reviewer perceived risk by generation strategy. The left panel shows expert methodological quality, calculated from expert ratings of research design, analysis strategy, validation, feasibility, and workflow integration. The right panel shows non-expert reviewer perceived risk; higher scores indicate greater perceived methodological or interpretive risk.}
\label{fig:secondary-outcomes}
\end{figure}

Expert methodological quality followed the same general direction as expert overall quality, whereas non-expert reviewer perceived risk was more variable and did not show a uniform directional reduction with skill augmentation (Figure~\ref{fig:secondary-outcomes}).

\subsection{Reviewer Agreement}

Expert agreement was limited in this small sample. For expert overall quality, the two-way absolute-agreement ICC was -0.15 for a single rating and -0.36 for the average rating; the mean absolute difference between the two expert ratings was 0.67 points. This mean expert disagreement was larger than the observed skill-minus-native expert overall-quality difference of 0.39 points. For expert methodological quality, the corresponding ICCs were 0.02 and 0.03, with a mean absolute difference of 0.54 points. Negative ICC estimates can occur in small samples when between-output variance is low relative to rater disagreement; therefore, expert mean ratings were interpreted descriptively rather than as stable latent quality estimates. Non-expert agreement was also modest using exploratory one-way random-effects ICCs because the non-expert design was not fully crossed: ICCs were 0.11 for quality, 0.03 for workflow, and 0.45 for perceived risk. These agreement results indicate substantial measurement noise and support treating the findings as descriptive and exploratory.

\subsection{Descriptive Model-Specific Effects of Skill Augmentation}

Skill effects varied by model, but these model-level values were not inferential estimates. GPT-5.4 showed the largest descriptive expert-recognized improvement (+1.25), followed by GLM-5.1 (+0.50). DeepSeek-V4 Pro and MiniMax-M2.7 showed mild descriptive improvements (+0.25 each). Kimi K2.6 was expert-neutral but strongly favored by non-expert reviewers (non-expert difference=+1.68). Claude Sonnet 4.6 showed no benefit and a mild decline (expert difference=-0.25; non-expert difference=-0.38). Model-level comparisons were based on one or two native-AI outputs and two skill-augmented outputs per model, so these estimates are best interpreted as signals for replication rather than evidence of model-specific performance.

\begin{table}[htbp]
\centering
\caption{Descriptive model-specific skill-minus-native differences.}
\scriptsize
\begin{tabularx}{\linewidth}{Y C{0.9cm} C{0.9cm} C{1.7cm} C{1.8cm} Y}
\toprule
Model & Native n & Skill n & Expert overall difference & Non-expert quality difference & Interpretation \\
\midrule
GPT-5.4 & 2 & 2 & +1.25 & +0.23 & Largest descriptive expert-recognized improvement \\
Claude Sonnet 4.6 & 1 & 2 & -0.25 & -0.38 & Mild decline or no benefit \\
GLM-5.1 & 1 & 2 & +0.50 & -0.17 & Expert-recognized improvement, non-expert neutral-to-negative \\
DeepSeek-V4 Pro & 2 & 2 & +0.25 & +0.13 & Mild descriptive improvement \\
Kimi K2.6 & 1 & 2 & 0.00 & +1.68 & Expert-neutral but non-expert favored \\
MiniMax-M2.7 & 2 & 2 & +0.25 & +0.30 & Mild descriptive improvement \\
\bottomrule
\end{tabularx}
\end{table}

\begin{figure}[htbp]
\centering
\includegraphics[width=0.95\linewidth]{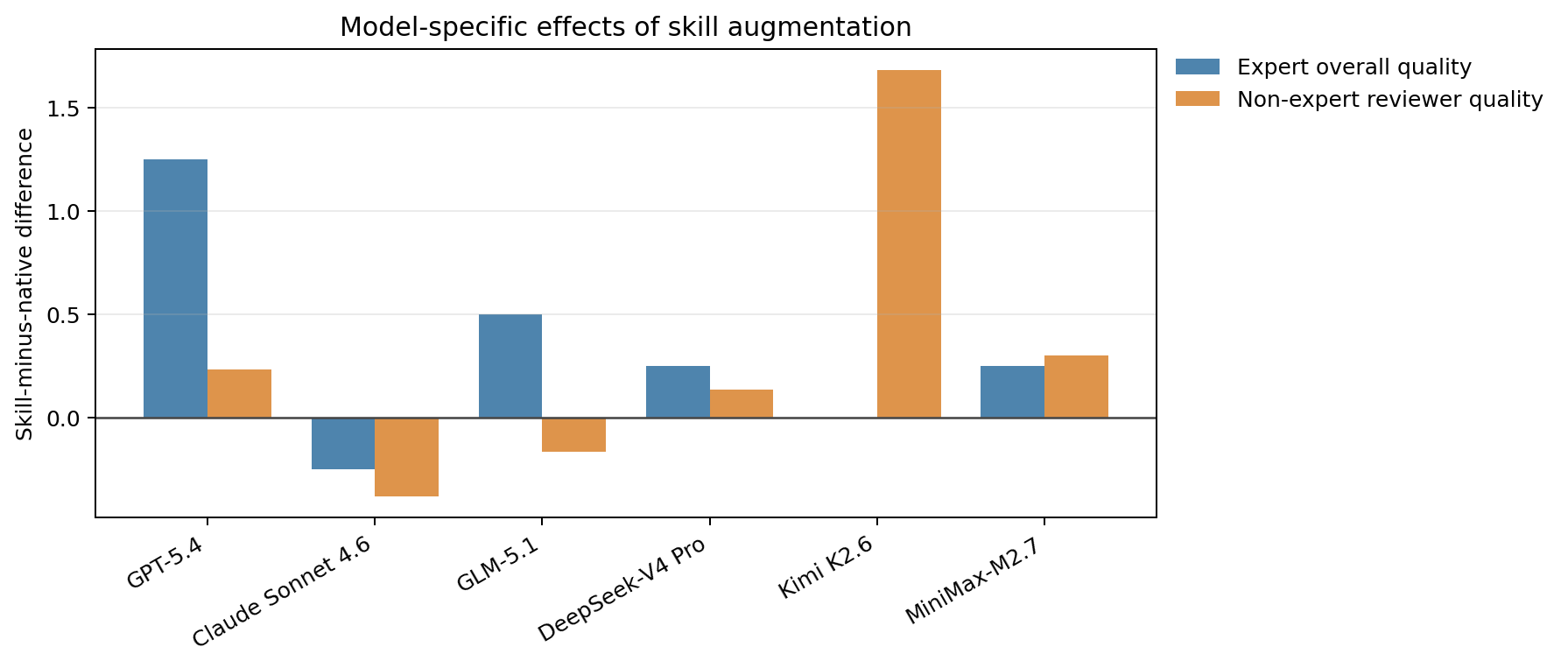}
\caption{Descriptive model-specific skill-minus-native differences. Bars show the model-level difference between skill-augmented and native-AI outputs for expert overall quality and non-expert reviewer quality. Positive values indicate higher scores for skill-augmented outputs, whereas negative values indicate lower scores after skill augmentation. These values are descriptive and not powered for model-level inference.}
\label{fig:model-specific-effects}
\end{figure}

\section{Discussion}

\subsection{Principal Findings}

This exploratory evaluation found a directional quality signal for AI-generated research-analysis outputs created with autonomous access to a medical research skill package. Skill-augmented outputs had higher mean expert overall quality and higher mean non-expert reviewer quality than native-AI outputs, but the confidence intervals crossed zero and expert agreement was limited. The expert overall-quality difference was smaller than the mean absolute disagreement between the two experts, so the primary result should be interpreted as a hypothesis-generating signal rather than evidence that skill augmentation reliably improves biomedical research-analysis quality. The effect was not uniform across models. GPT-5.4 and GLM-5.1 showed the largest descriptive expert-recognized gains; DeepSeek-V4 Pro and MiniMax-M2.7 showed mild gains; Kimi K2.6 showed non-expert-perceived improvement without expert-rated improvement; and Claude Sonnet 4.6 showed no benefit. This pattern is consistent with prior work emphasizing that tool- or skill-augmented agents should be evaluated at the workflow level rather than only by language quality [3-7,11].

These findings should therefore be interpreted at two levels. At the generation-strategy level, skill augmentation was associated with higher mean quality scores than native AI, but the evidence remains exploratory and measurement-limited. At the model level, the magnitude and direction of the effect differed substantially across model backbones, indicating that the benefit of skill access was not model-independent and cannot be separated from small cell sizes.

\subsection{Interpretation}

The findings suggest that skill augmentation may be most useful when native-AI outputs lack structure, validation logic, or methodological completeness. The large descriptive GPT-5.4 difference suggests that the skill package may compensate for a weaker native baseline by improving method selection, validation design, workflow integration, and feasibility. GLM-5.1 also showed a positive expert-rated difference, suggesting that some improvements were methodological and therefore more visible to experts than to non-expert reviewers. This interpretation aligns with agent skill benchmarks that frame skills as reusable procedural units whose value depends on task fit and successful execution [7,11].

Kimi K2.6 illustrates a different mechanism: skill access was not associated with higher expert overall quality but was associated with substantially higher non-expert reviewer perception. This suggests that the skill package may have improved readability, organization, and apparent usability, while expert-level concerns about evidence quality, endpoint consistency, or validation remained. Claude Sonnet 4.6 showed the opposite pattern. Its native baseline was already strong, and additional skill-driven process detail may have increased complexity without improving expert confidence.

\subsection{Measurement Validity and Human Evaluation Noise}

The most important methodological finding may be the limited reliability of expert judgments in this setting. Expert overall-quality ICC was negative, and the observed expert skill-minus-native difference was smaller than the mean absolute disagreement between experts rating the same output. This does not invalidate the descriptive comparison, but it does mean that the primary outcome should not be treated as a precise latent measure of output quality. The low agreement may reflect limitations of the rating instrument, insufficient rater calibration, genuine ambiguity in evaluating complex AI-generated research plans, or low between-output variance relative to reviewer judgment. Future evaluations should include more experts, explicit rater training, calibration examples, adjudication or consensus review, and possibly separate scoring of factual correctness, methodological appropriateness, biological plausibility, and workflow coherence.

\subsection{Biological Validity Gap}

This study evaluated human-rated research-analysis quality rather than the biological validity of proposed biomarkers. A high-scoring output may be clearer, more complete, and more methodologically structured while still proposing a biologically weak gene signature or overinterpreting transcriptomic associations. Conversely, a less polished output could contain a plausible biological lead that is not reflected in a generic quality score. The PCD component of the benchmark should therefore be interpreted as a mechanistic stress test built on broad bioinformatics modules, not as proof that the agent had disease- and mechanism-specific expertise. The workflow did include a dataset-relevance check to ensure that downloaded public data were lung-cancer or NSCLC related. However, we did not independently validate dedicated ferroptosis, cuproptosis, or pyroptosis knowledge modules, nor did we audit whether outputs used authoritative PCD gene sets rather than model-generated or loosely cited gene lists. For NSCLC immunotherapy research, biological validity would require independent verification of cohort labels, endpoint consistency, mechanism-specific PCD gene sets, feature-selection stability, external validation, and alignment with known immunotherapy biology. Therefore, the present results should be interpreted as evidence about AI-assisted research-analysis support, not as evidence that skill-augmented agents produce clinically valid biomarkers.

\subsection{Possible Mechanisms Underlying Model-Specific Differences}

The heterogeneous effects observed across model backbones may be explained by differences in baseline model capability, orchestration behavior, and skill comprehension. First, models differed in their native ability to decompose a biomedical research task into coherent analytical subtasks. A model with weaker native planning may benefit substantially from explicit skill access because the skills provide a scaffold for dataset selection, preprocessing, endpoint definition, biomarker modeling, validation, and biological interpretation. In contrast, a model that already produces a coherent native research workflow may gain less from additional skills, and may even lose quality if skill invocation adds redundant steps, excessive procedural detail, or conflicts with the model's own higher-level reasoning. This supports the view that skill systems are not independent replacements for model reasoning but interact with the underlying model's planning capacity [7,10,11].

Second, skill augmentation depends not only on whether relevant skills are available, but also on how the agent routes and sequences them. The same skill package can produce different outputs if the model selects too many skills, omits upstream dependencies, invokes methods before required inputs are available, or treats optional downstream analyses as mandatory. In transcriptomic biomarker research, early orchestration decisions are especially consequential: incorrect cohort selection, endpoint mapping, or expression matrix handling can propagate into feature selection, ROC analysis, immune infiltration analysis, and mechanistic interpretation. Skill orchestration quality is therefore a plausible mediator between skill availability and final output quality, consistent with recent work on skill routing, dependency-aware skill retrieval, and scalable skill-flow systems [8,9,12].

Third, models may differ in how well they understand the intended function, assumptions, and boundaries of each skill. Strong skill comprehension requires recognizing what a skill is designed to do, what input it requires, what output should be passed downstream, and when the skill should be skipped or flagged as infeasible. Some models may treat skills as conceptual prompts and generate plausible descriptions without fully respecting execution constraints. Others may better preserve input-output continuity but still over-apply methods when sample size, endpoint availability, or validation data are insufficient. This distinction may explain why some skill-augmented outputs appeared clearer to non-expert reviewers but did not consistently improve expert-rated credibility. Similar concerns have been raised in tool-use benchmarks, where access to tools or APIs does not guarantee correct tool selection, argument construction, or result interpretation [3-6].

Fourth, models may vary in their ability to integrate multiple skill outputs into a unified research analysis. A high-quality output should not merely concatenate evidence search, protocol design, differential expression analysis, immune analysis, and modeling sections. It should preserve a coherent methodological chain: the research gap should motivate the cohort design; the endpoint should determine the modeling strategy; the preprocessing pipeline should support valid downstream comparisons; and mechanistic analyses should be aligned with the final biomarker claim. When this integration is incomplete, skill use can increase apparent completeness while leaving unresolved methodological inconsistencies.

\subsection{Sources of Score Variability}

Score variability likely arose from several factors. First, baseline model capability differed, so the marginal value of skills was higher for weaker or less structured native outputs. Second, agentic outputs showed run-to-run variability: early decisions about dataset selection, endpoint interpretation, validation feasibility, and whether a step was executed or merely described could propagate downstream. Third, non-expert reviewers and experts emphasized different dimensions. Non-expert reviewers tended to reward clarity and usability, whereas experts weighted endpoint validity, sample size, overfitting risk, statistical assumptions, and claim boundaries more heavily. Fourth, inter-rater agreement was limited, particularly for expert ratings, indicating that the scoring task itself involved substantial judgment. Finally, full generated materials were preserved during review, including logs, tables, figures, scripts, and intermediate results; fragmented or inconsistent artifacts could reduce expert confidence even when the narrative appeared complete.

\subsection{Implications}

These results support medical research skill packages as a promising strategy for improving AI-generated research-analysis outputs, but they do not establish a uniform or confirmatory performance advantage. The heterogeneous model-specific effects suggest that skill augmentation should be adaptive. Future agents should not simply expose more skills; they should route skills according to task requirements, data availability, endpoint structure, sample size, and validation feasibility. They should also explicitly document skipped steps, failed analyses, claim boundaries, and required human review. These design requirements are consistent with emerging skill orchestration frameworks that emphasize routing, dependency structure, benchmarking, and workflow-level evaluation [7-12].

The study also suggests that future benchmarks should include stronger controls. An irrelevant-skill or placebo-skill condition would help distinguish genuine skill value from the effect of simply providing additional structured context. Platform-agnostic replications would help separate the contribution of the medical research skill package from the execution behavior of a particular AI agent platform. Finally, downstream validation tasks should test whether improved perceived quality leads to better dataset selection, fewer methodological errors, and more biologically defensible biomarker hypotheses.

\subsection{Limitations}

This study was exploratory and small. The evaluation dataset included 21 outputs across six model backbones, with repeated runs retained for some model-strategy combinations. Statistical tests should therefore be interpreted as hypothesis-generating rather than confirmatory. Repeated outputs and ratings were summarized at the output and model levels, but the design was not powered for definitive mixed-effects inference. No multiplicity adjustment was applied, so favorable secondary or model-level contrasts may reflect false-positive or selection effects. Reviewer agreement was limited, the rating instrument was developed for this study rather than externally validated, and non-expert ratings were not fully crossed by reviewer. The study evaluated one biomedical research scenario and may not generalize to other diseases, modalities, or tasks. OpenClaw was used as the representative AI agent platform, so the study cannot separate the effects of the skill package from OpenClaw-specific routing, execution, context management, or artifact-generation behavior. The review assessed perceived quality of AI-generated research assistance, not the clinical validity of any proposed biomarker. The model set was selected pragmatically and does not constitute a systematic or exhaustive benchmark of current LLMs.

\section{Conclusion}

In this multi-model evaluation, autonomous access to a medical research skill package produced a directional but non-confirmatory quality signal for AI-generated transcriptomic research-analysis outputs. Descriptive model-level differences suggested heterogeneous patterns across model backbones, but the small cell sizes and limited expert agreement prevent model-level inference. The main value of this study is to define a biomedical workflow-level evaluation setting and to show that future skill-augmented AI agent studies need stronger reliability, platform, and biological-validity controls before performance claims can be made.

\section{Ethics and Availability Statements}

This preprint reports an evaluation of anonymized AI-generated research-analysis documents. Reviewers evaluated document quality and did not provide patient-level data, clinical information, biological samples, or intervention-related outcomes. Reviewers participated voluntarily in an internal document-evaluation process. The study did not collect patient or public health data and did not involve clinical intervention, patient contact, biological specimens, or treatment-related decision-making. No institution-specific ethics approval number is reported for this preprint.

The medical research skill package is publicly available in the \href{https://github.com/aipoch/medical-research-skills/tree/main/awesome-med-research-skills}{AIPOCH medical research skills repository}. The arXiv source package includes the reproducibility supplement, aggregate analysis CSV files, and figures needed to reproduce the reported summary tables and plots. The complete generated-output archive is not deposited with this preprint because it contains operational metadata, internal file paths, and platform-specific traces that require additional de-identification. De-identified generated materials may be made available by the authors upon reasonable request after removal of residual operational metadata and any information that could compromise review blinding or internal system configuration.

\section*{References}
\begin{enumerate}

\item Singhal, K., Azizi, S., Tu, T., et al. (2023). Large language models encode clinical knowledge. \textit{Nature, 620}, 172-180. \url{https://www.nature.com/articles/s41586-023-06291-2}

\item Moor, M., Banerjee, O., Abad, Z. S. H., et al. (2023). Foundation models for generalist medical artificial intelligence. \textit{Nature, 616}, 259-265. \url{https://www.nature.com/articles/s41586-023-05881-4}

\item Li, M., Song, F., Yu, B., et al. (2023). API-Bank: A comprehensive benchmark for tool-augmented LLMs. \textit{arXiv}. \url{https://arxiv.org/abs/2304.08244}

\item Patil, S. G., Zhang, T., Wang, X., \& Gonzalez, J. E. (2023). Gorilla: Large language model connected with massive APIs. \textit{arXiv}. \url{https://arxiv.org/abs/2305.15334}

\item Qin, Y., Liang, S., Ye, Y., et al. (2023). ToolLLM: Facilitating large language models to master 16000+ real-world APIs. \textit{arXiv}. \url{https://arxiv.org/abs/2307.16789}

\item Shen, Y., Song, K., Tan, X., et al. (2023). TaskBench: Benchmarking large language models for task automation. \textit{arXiv}. \url{https://arxiv.org/abs/2311.18760}

\item Li, H., Mu, C., Chen, J., Ren, S., Cui, Z., Zhang, Y., Bai, L., Hu, S., et al. (2026). AgentSkillOS: Organizing, orchestrating, and benchmarking agent skills at ecosystem scale. \textit{arXiv}. \url{https://arxiv.org/abs/2603.02176}

\item Zheng, Y., Zhang, Z., Ma, C., Yu, Y., Zhu, J., Dong, B., \& Zhu, H. (2026). SkillRouter: Skill routing for LLM agents at scale. \textit{arXiv}. \url{https://arxiv.org/abs/2603.22455}

\item Li, D., Li, Z., Du, H., Wu, X., Gui, S., Kuang, Y., \& Sun, L. (2026). Graph of Skills: Dependency-aware structural retrieval for massive agent skills. \textit{arXiv}. \url{https://arxiv.org/abs/2604.05333}

\item Wang, J., Ming, Y., Ke, Z., Joty, S., Albarghouthi, A., \& Sala, F. (2026). SkillOrchestra: Learning to route agents via skill transfer. \textit{arXiv}. \url{https://arxiv.org/abs/2602.19672}

\item Li, X., Chen, W., Liu, Y., Zheng, S., Chen, X., He, Y., Li, Y., You, B., Shen, H., Sun, J., Wang, S., Zeng, Q., Wang, D., Zhao, X., Wang, Y., Ben Chaim, R., Di, Z., Gao, Y., He, J., et al. (2026). SkillsBench: Benchmarking how well agent skills work across diverse tasks. \textit{arXiv}. \url{https://arxiv.org/abs/2602.12670}

\item Li, F., Tagkopoulos, P., \& Tagkopoulos, I. (2025). SkillFlow: Scalable and efficient agent skill retrieval system. \textit{arXiv}. \url{https://arxiv.org/abs/2504.06188}
\end{enumerate}
\clearpage
\appendix
\input{supplement.tex}
\end{document}

%% file: supplement.tex
\setcounter{table}{0}
\renewcommand{\thetable}{S\arabic{table}}

\section{Supplementary Reproducibility Information}

This supplement provides reproducibility details that are not fully expanded in the main text, including the exact task prompt, output inclusion rules, evaluated output counts, rating anchors, analysis artifacts, and reproducibility boundaries.

\section{S1. Unified Task Prompt}

All models were asked to complete the same biomedical research-analysis task:

\begin{verbatim}
Based on public transcriptomic data, construct a multi-gene signature
for predicting immunotherapy response in non-small cell lung cancer
(NSCLC), and explore the role of programmed cell death mechanisms,
including ferroptosis, cuproptosis, and pyroptosis, in immunotherapy
resistance.

Please output a research-analysis plan and complete data-analysis
workflow with a level of complexity comparable to a journal article
with an impact factor (IF) of approximately 5 in the corresponding year.

The content should cover public dataset selection, cohort design,
endpoint definition, data preprocessing, differential expression
analysis, candidate gene screening, model construction, validation
strategy, immune microenvironment analysis, mechanistic interpretation,
key figures and tables, and manual review points.
\end{verbatim}

For the skill-augmented strategy, the same task was executed in an artificial intelligence (AI) agent environment with autonomous access to the medical research skill package. The agent was allowed to select relevant skills without a fixed human-imposed sequence.

\section{S2. Output Inclusion Rules and Evaluated Output Counts}

Outputs were retained if they produced reviewable research-analysis material, including:

\begin{itemize}
\item complete research-analysis outputs;
\item partial research reports with sufficient task-relevant content;
\item stage summaries;
\item archived traces that contained sufficient task-relevant content for human review.
\end{itemize}

Outputs were not excluded solely because the final artifact format differed across model or agent runs, provided that the archived material contained enough content for blinded human review. This rule was used to preserve real-world agentic-output variability.

\begin{table}[htbp]
\centering
\caption{Evaluated outputs by model and generation strategy.}
\label{tab:supp-output-counts}
\small
\begin{tabularx}{\linewidth}{Y C{1.8cm} C{1.8cm} Y}
\toprule
Model & Native-AI outputs & Skill-augmented outputs & Notes \\
\midrule
GPT-5.4 & 2 & 2 & Included in model-specific descriptive analysis \\
Claude Sonnet 4.6 & 1 & 2 & Included in model-specific descriptive analysis \\
GLM-5.1 & 1 & 2 & Included in model-specific descriptive analysis \\
DeepSeek-V4 Pro & 2 & 2 & Included in model-specific descriptive analysis \\
Kimi K2.6 & 1 & 2 & Included in model-specific descriptive analysis \\
MiniMax-M2.7 & 2 & 2 & Included in model-specific descriptive analysis \\
\bottomrule
\end{tabularx}
\end{table}

The final analysis included 21 anonymized outputs in total: 9 native-AI outputs and 12 skill-augmented outputs. Anonymous output identifiers were used during blinded review, but the complete identifier-to-output archive is not included in the arXiv source package because the full generated materials require additional de-identification before public release.

\section{S3. Rating Scale Anchors and Reviewer Design}

All Likert-scale items used a 1-7 scale. Each anonymized output received two non-expert biomedical reviewer ratings and two blinded expert ratings. Reviewers did not receive model names, generation strategies, platform configuration, or the mapping between anonymous output IDs and experimental conditions.

\begin{table}[htbp]
\centering
\caption{Rating scale anchors.}
\label{tab:supp-rating-anchors}
\small
\begin{tabularx}{\linewidth}{Y Y Y Y}
\toprule
Construct & Anchor 1 & Anchor 4 & Anchor 7 \\
\midrule
Quality, clarity, completeness, feasibility, usability & Very poor / strongly disagree & Neutral or partially acceptable & Excellent / strongly agree \\
Workflow coherence & Incoherent or poorly integrated & Partially coherent & Highly coherent and well integrated \\
Expert methodological quality & Methodologically inappropriate & Partially appropriate with limitations & Methodologically strong and appropriate \\
Perceived risk & Very low perceived risk & Moderate or uncertain risk & Very high perceived risk \\
\bottomrule
\end{tabularx}
\end{table}

For quality and workflow constructs, higher scores indicate better evaluation. For perceived-risk constructs, higher scores indicate greater perceived methodological or interpretive risk.

\section{S4. Displayed Analysis Artifacts}

The following analysis artifacts are displayed in the compiled arXiv PDF. Aggregate CSV files used for analysis are retained in the source package for reproducibility but are not listed here as display artifacts because CSV files are not rendered as manuscript content during submission.

\begin{table}[htbp]
\centering
\caption{Displayed analysis artifacts.}
\label{tab:supp-analysis-artifacts}
\small
\begin{tabularx}{\linewidth}{Y Y}
\toprule
File & Purpose \\
\midrule
\path{figure0_quality_control_flow.png} & Evaluation dataset quality-control flow figure \\
\path{figure1_overall_quality_by_strategy.png} & Overall expert and non-expert quality figure \\
\path{figure2_secondary_outcomes_by_strategy.png} & Expert methodological quality and perceived risk figure \\
\path{figure3_model_specific_skill_effects.png} & Model-specific skill-minus-native effects figure \\
\bottomrule
\end{tabularx}
\end{table}

\section{S5. Reproducibility Boundaries}

The study is reproducible at the level of task prompt, model labels, strategy definitions, output inclusion criteria, anonymized output IDs, rating constructs, and analysis outputs. It is not claimed to be exactly reproducible at the level of proprietary model sampling internals, hidden provider-side model revisions, or OpenClaw deployment internals that were not exported in the available run archives.